\PassOptionsToPackage{square,numbers}{natbib}
\PassOptionsToPackage{hyphens}{url}

\documentclass{article}

     \usepackage[preprint]{neurips_2019}

\usepackage[utf8]{inputenc} %
\usepackage[T1]{fontenc}    %
\usepackage{hyperref}       %
\usepackage{url}            %
\usepackage{booktabs}       %
\usepackage{amsfonts}       %
\usepackage{nicefrac}       %
\usepackage{microtype}      %

\usepackage{etoolbox}
\usepackage{mathtools}
\usepackage{amssymb}
\usepackage{multirow}
\usepackage{xcolor}
\usepackage{subcaption}
\usepackage{hyperref}
\hypersetup{breaklinks=true}

\newtoggle{include-notes}
\togglefalse{include-notes}

\allowdisplaybreaks

\newcommand\bestresult{42.94\%}

\title{Encoding Database Schemas with Relation-Aware Self-Attention for Text-to-SQL Parsers}

\author{%
  Richard Shin\thanks{Work done partly during an internship at Intel Labs.} \\
  UC Berkeley \\
  \texttt{ricshin@cs.berkeley.edu} \\
}

\begin{document}

\maketitle

\begin{abstract}
  When translating natural language questions into SQL queries to answer questions from a database,
  we would like our methods to generalize to domains and database schemas outside of the training set.
  To handle complex questions and database schemas with a neural encoder-decoder paradigm, it is critical to properly encode the schema as part of the input with the question.
  In this paper, we use relation-aware self-attention within the encoder 
  so that it can reason about how the tables and columns in the provided schema relate to each other
  and use this information in interpreting the question.
  We achieve significant gains on the recently-released Spider dataset
  with \bestresult{} exact match accuracy, compared to the $18.96\%$ reported in published work.
\end{abstract}

\section{Introduction}
\label{sec:introduction}
The ability to effectively query databases with natural language has the potential to unlock the power of large datasets to the vast majority of users who are not proficient in the use of languages such as SQL.
As such, a large body of existing work has focused on the task of translating natural language questions into queries that existing database software can execute.

The release of large annotated datasets containing questions and the corresponding database queries has catalyzed significant progress in the field, by enabling the training of supervised learning models for the task \citep{zhongSeq2SQLGeneratingStructured2017,finegan-dollakImprovingTexttoSQLEvaluation2018}.
This progress has arrived not only in the form of improved accuracy on the test sets provided with the datasets, but also through an evolution of the problem formulation towards greater complexity more closely resembling real-world applications.

The recently-released Spider dataset \citep{data-spider} exemplifies greater realism in the task specification: the queries are written using SQL syntax, the dataset contains a large number of domains and schemas with no overlap between the train and test sets, and each schema contains multiple tables with many complicated questions being expressed in the queries.
Due to the extra difficulty caused by these factors, the best result on this dataset in published work achieves about 19\% exact match accuracy on the development set \citep{yuSyntaxSQLNetSyntaxTree2018},
which is significantly worse compared to $>80\%$ exact matching accuracy reported for past datasets such as ATIS, GeoQuery, and WikiSQL \citep{data-spider,LargeAnnotatedSemantic2019}.

\begin{figure}[t]
    \centering
    \includegraphics[width=0.9\columnwidth]{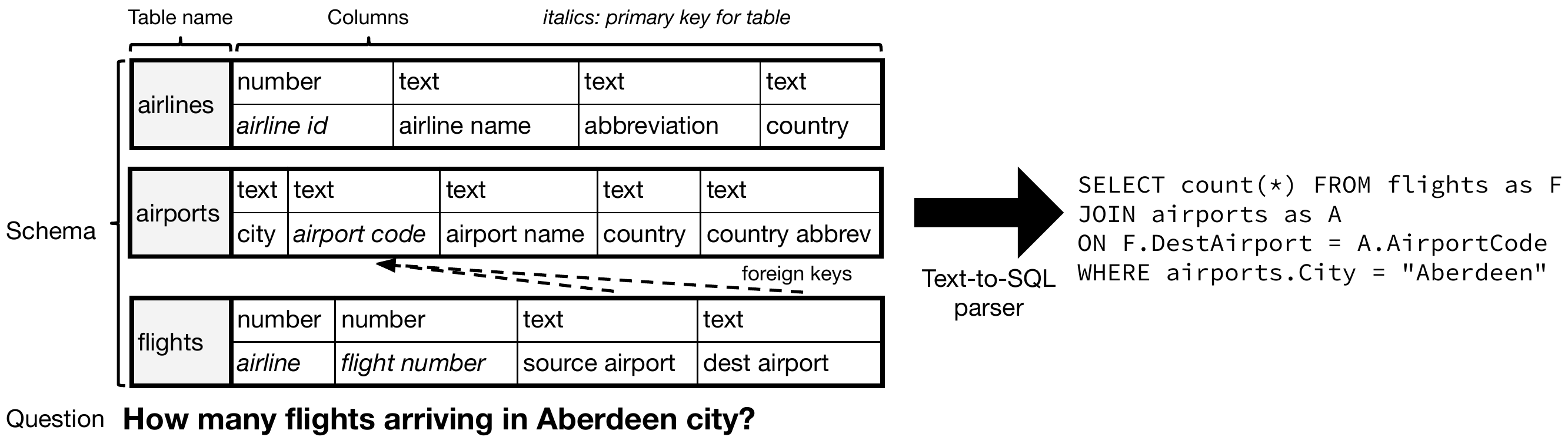}

    \caption{Overview of text-to-SQL task.
    This paper proposes and evaluates the use of relation-aware self-attention to encode the question and schema, including elements such as the ``foreign key'' relationship shown.}
    \label{fig:task-overview}
\end{figure}

We posit that a central challenge of the multi-schema problem setting is generalization to new database schemas different from what was seen during training.
When the model needs to generate queries for arbitrary new schemas, it needs to take the relevant schema as an input and process it together with the question in order to generate the correct query.

Previous methods on the WikiSQL dataset \citep{zhongSeq2SQLGeneratingStructured2017} have also contended with the challenge of generalizing to arbitrary new schemas.
However, all schemas in this dataset are quite simple, as they only contain one table.
The model has no need to reason about the relationships between multiple tables in order to generate the correct query.
As such, models developed for this dataset have largely focused on innovations to the decoder for generating the query,
rather than the encoder for the question and the schema. 
In contrast, most real databases (including those in Spider) contain multiple tables with features such as foreign keys that link rows in one table to another. 
We hypothesize that to generate correct queries for such databases,
a model needs the ability to reason about how the tables and columns in the provided schema relate to each other
and use this information in interpreting the question.

In this paper, we develop a method to test this hypothesis.
First, we construct a directed graph (with labels on nodes and edges) over all of the elements of the schema.
This graph contains a node for each column or table, and an edge exists from one node to another if the two have an interesting relationship (e.g., the two nodes are columns which belong to the same table) with a label encoding that relationship.
Each node has an initial vector representation based on the words in the column or table's name.
We also obtain a vector representation for each word in the question.
For a fixed number of times, we then update each node and word representation based on all other node and word representations, taking the labels of edges between nodes into account. %
We use these updated representations with a tree-structured SQL decoder, which uses attention over them at each decoding step, and also points to the column and table representations when it needs to output a column or table reference in the query.

We empirically evaluate our method on the Spider dataset~\citep{data-spider}, using a decoder based on \citet{yinSyntacticNeuralModel2017a}.
We achieve \bestresult{} exact set match accuracy on the development set, significantly higher than the published result of 18.9\%~\citep{yuSyntaxSQLNetSyntaxTree2018}.
We further verify the utility of directly encoding the relationships within the schema with an ablation study.
\section{Problem Formulation and Motivation}
Provided with a natural language question and a schema for a relational database, our goal is to generate the SQL query corresponding to the question.
The schema contains the following information, as depicted in Figure~\ref{fig:task-overview}:
a list of \emph{tables} in the database, each with a meaningful name (e.g., \textsc{airlines}, \textsc{airports}, and \textsc{flights} for an aviation database);
for each table, a list of \emph{columns}, where
each column has a type such as \texttt{number} or \texttt{text},
and some of them can be \emph{primary keys}, used to uniquely identify each row;
finally, a column can have another column in a different table as its \emph{foreign key}, which is used to link together rows across multiple tables.
As mentioned in the introduction, we would like our method to generalize to not only new questions, but also new schemas it has never seen during training time.
\label{sec:motivation}
Using natural language to query databases has been a long-standing problem studied for many decades in the research community
\citep{androutsopoulosNaturalLanguageInterfaces1995,popescuModernNaturalLanguage2004}.
We identify several limitations of past work and problem settings:%

\renewcommand{\labelenumi}{(\alph{enumi})}
\begin{enumerate}
  \item Some datasets only concern themselves with one domain (e.g., US geography \citep{data-geography-original}).
  \item Most datasets about one domain also contain only one database schema for the domain, so the system only needs to know how to generate queries for that single schema.
  \item While WikiSQL \citep{zhongSeq2SQLGeneratingStructured2017} contains a large number of domains and schemas, each schema only contains one table in it.%
  \item Datasets containing only one domain and database necessarily contain overlaps between the train and test sets. Furthermore, as discussed by \citet{finegan-dollakImprovingTexttoSQLEvaluation2018}, many existing datasets exhibit overlap in queries between the train and test sets, which limits their ability to test how models generalize to generating new queries.
\end{enumerate}

The neural methods common in recent work follow an encoder-decoder paradigm,
and
past work has largely focused on improvements to the decoder part.
As such, the question of how best to encode the question and the schema has remained relatively under-studied.
Models developed using datasets which contain only one domain and schema ((a) and (b) above) typically internalize the schema within the learned parameters.
The popular WikiSQL dataset necessitates generalizing to new schemas at test time, so models developed for it also encode the schema together with the question;
however, as all these schemas only contain one table, the demands placed on the schema encoder are relatively light.

It is most useful if we can train a single model that can generalize to new domains and new database schemas,
where both the queries and the schemas have complicated structure that better reflect potential real-world applications.
The Spider dataset \citep{data-spider} provides an environment for evaluating this problem setting.
In this work, we study how to better encode the question and schema under these more demanding conditions.

\section{Existing Encoding Schemes}
In this section, we review how some existing works (mostly for the WikiSQL dataset) addressed the challenge of encoding the input question and schema.

\paragraph{Encoding each element independently}
In SQLNet \citep{xuSQLNetGeneratingStructured2017} (for the WikiSQL dataset),
the name of each column, and the question, are separately processed using a bidirectional LSTM.
The LSTM outputs for the question tokens are utilized in the decoder using attention, and the final LSTM states of the columns with a pointer network.
Note that the encoding of each column is uninfluenced by which other column are present;
furthermore, the question is encoded entirely separately from the schema.

In SyntaxSQLNet \citep{yuSyntaxSQLNetSyntaxTree2018} (for the Spider dataset),
the question is encoded identically as SQLNet, using a bidirectional LSTM.
Each column is encoded similarly, by using a bidirectional LSTM over the concatenation of the words in the column name, words in the table name, and column type (e.g., \texttt{number}, \texttt{string}).

\paragraph{Encoding the columns jointly}
TypeSQL \citep{yuTypeSQLKnowledgeBasedTypeAware2018} computes the encoding of each column by an elementwise averaging of the embeddings of the words in the name, and using a bidirectional LSTM over these averages (i.e., over all columns);
therefore, the encoding for each column depends on which other columns are present (and also their order, although that is arbitrary).

\paragraph{Using the schema while encoding the question}
Using the information in the schema while encoding the question can help the decoder generate the correct query.
In TypeSQL, the word embeddings for each question token are concatenated with a \emph{type} embedding;
in particular, question tokens appearing in a column name are specially marked.

Coarse2Fine \citep{dongCoarsetoFineDecodingNeural2018} goes further
by using attention to gather information from the schema while encoding the question.
First, the input question is encoded using a bidirectional LSTM,
then an attention mechanism retrieves a weighted sum of the column embeddings for the LSTM state of each token.
These two are concatenated together and processed together in another bidirectional LSTM, to obtain the final embeddings for each question token.

IncSQL \citep{shiIncSQLTrainingIncremental2018} uses ``cross-serial attention'', also updating the column embeddings using the question token embeddings, in addition to the other direction used in Coarse2Fine.
\section{Our Approach}
\label{sec:our-encoder}
In the previous section, we reviewed how previous neural methods developed for the text-to-SQL problem encode the input (the question and the database schema) for use in the decoder.
Several of these methods encode the question and the columns entirely independently (e.g., the embedding of a column is uninfluenced by other columns in the schema).

In contrast, we specifically seek interactions between schema elements within our encoder,
as explained in Sections \ref{sec:introduction} and \ref{sec:motivation}.
In this section, we describe how we encode the schema as a directed graph and use relation-aware self-attention to interpret it.
We will use the following notation:
\begin{itemize}
    \item $c_i$ for each column in the schema. Each column contains words $c_{i,1}, \cdots, c_{i,|c_i|}$.
    \item $t_i$ for each table in the schema. Each table contains words $t_{i,1}, \cdots, t_{i,|t_i|}$.
    \item $q$ for the input question. The question contains words $q_1, \cdots, q_{|q|}$.
\end{itemize}

\begin{table}[t]
\caption{Description of edge types present in the directed graph created to represent the schema. An edge exists from node $x$ to node $y$ if the pair fulfills one of the descriptions listed in the table, with the corresponding label. Otherwise, no edge exists from $x$ to $y$.}
\label{table:schema-graph-edges}
\centering
\begin{tabular}{lllp{6cm}}
\toprule
Type of $x$ & Type of $y$ & Edge label & Description \\
\midrule
\multirow{3}{*}{Column} & \multirow{3}{*}{Column}
   & \textsc{Same-Table}    & $x$ and $y$ belong to the same table. \\
 & & \textsc{Foreign-Key-Col-F} & $x$ is a foreign key for $y$. \\
 & & \textsc{Foreign-Key-Col-R} & $y$ is a foreign key for $x$. \\
\midrule
\multirow{2}{*}{Column} & \multirow{2}{*}{Table}
   & \textsc{Primary-Key-F}   & $x$ is the primary key of $y$. \\
 & & \textsc{Belongs-To-F}    & $x$ is a column of $y$ (but not the primary key). \\
\midrule
\multirow{2}{*}{Table} & \multirow{2}{*}{Column}
   & \textsc{Primary-Key-R}   & $y$ is the primary key of $x$. \\
 & & \textsc{Belongs-To-R}    & $y$ is a column of $x$ (but not the primary key). \\
\midrule
\multirow{3}{*}{Table} & \multirow{3}{*}{Table}
   & \textsc{Foreign-Key-Tab-F}   & Table $x$ has a foreign key column in $y$. \\
 & & \textsc{Foreign-Key-Tab-R}   & Same as above, but $x$ and $y$ are reversed. \\
 & & \textsc{Foreign-Key-Tab-B}   & $x$ and $y$ have foreign keys in both directions. \\
\bottomrule
\end{tabular}
\end{table}

\begin{figure}[t]
    \centering
    \includegraphics[width=0.7\columnwidth]{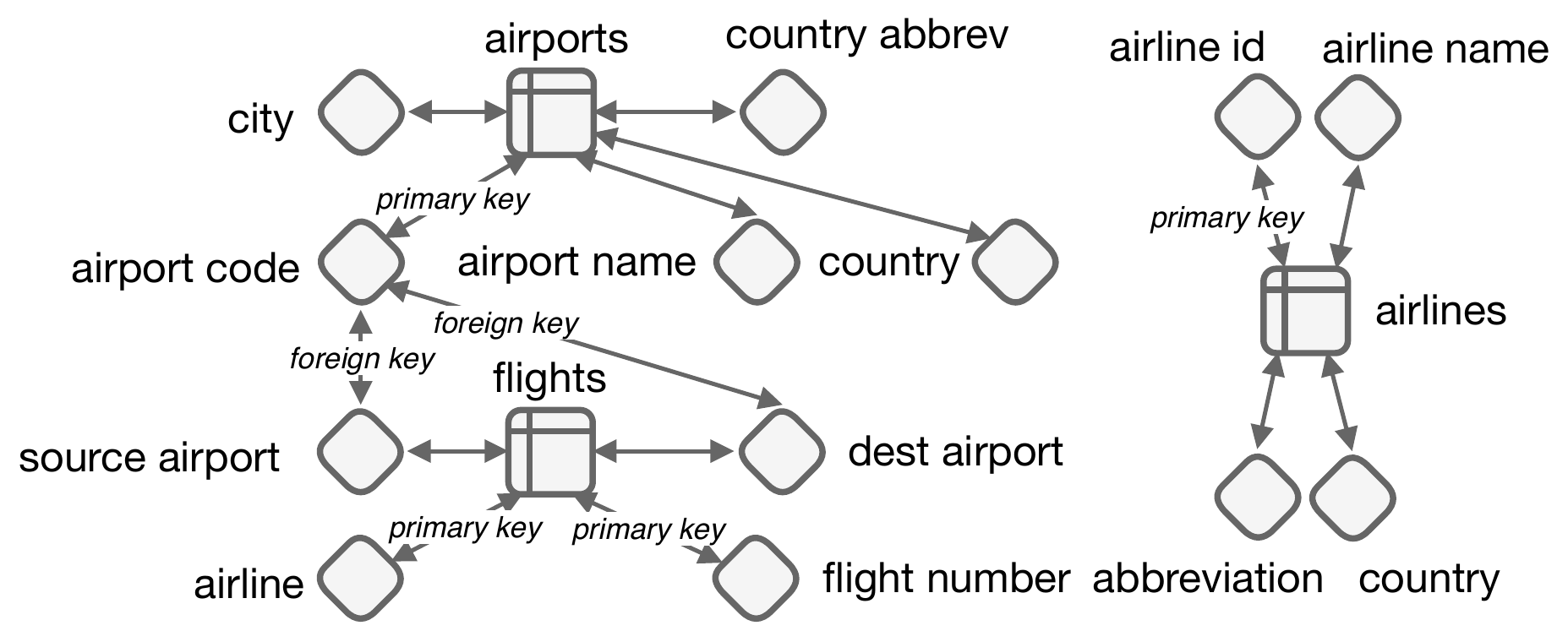}
    \caption{An illustration of an example schema as a graph. We do not depict all edges and label types of Table~\ref{table:schema-graph-edges} to reduce clutter.}
    \label{fig:schema-graph}
\end{figure}

\begin{figure}[t]
    \centering
    \begin{subfigure}[b]{0.305\textwidth}
        \includegraphics[width=\textwidth]{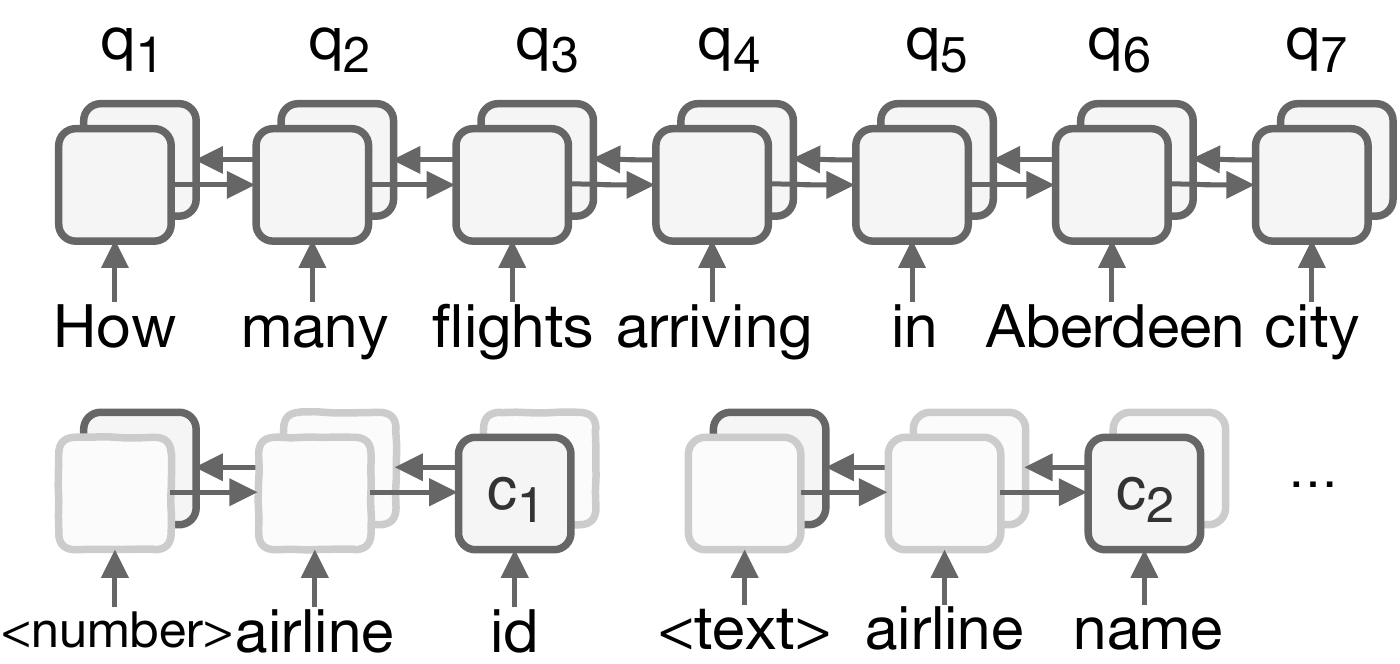}
        \caption{Initial encoding of the input (Section~\ref{sec:initial-encoding})}
        \label{fig:initial-encoding}
    \end{subfigure}
    ~%
    \begin{subfigure}[b]{0.305\textwidth}
        \includegraphics[width=\textwidth]{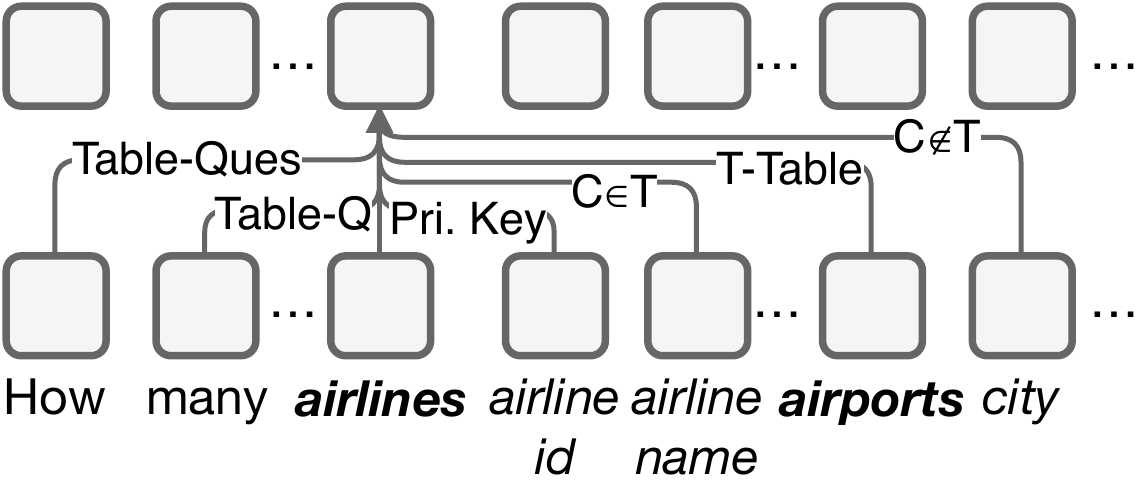}
        \caption{One layer of relation-aware self-attention (Section~\ref{sec:rel-attn})}
        \label{fig:rel-attn}
    \end{subfigure}
    ~%
    \begin{subfigure}[b]{0.305\textwidth}
        \includegraphics[width=\textwidth]{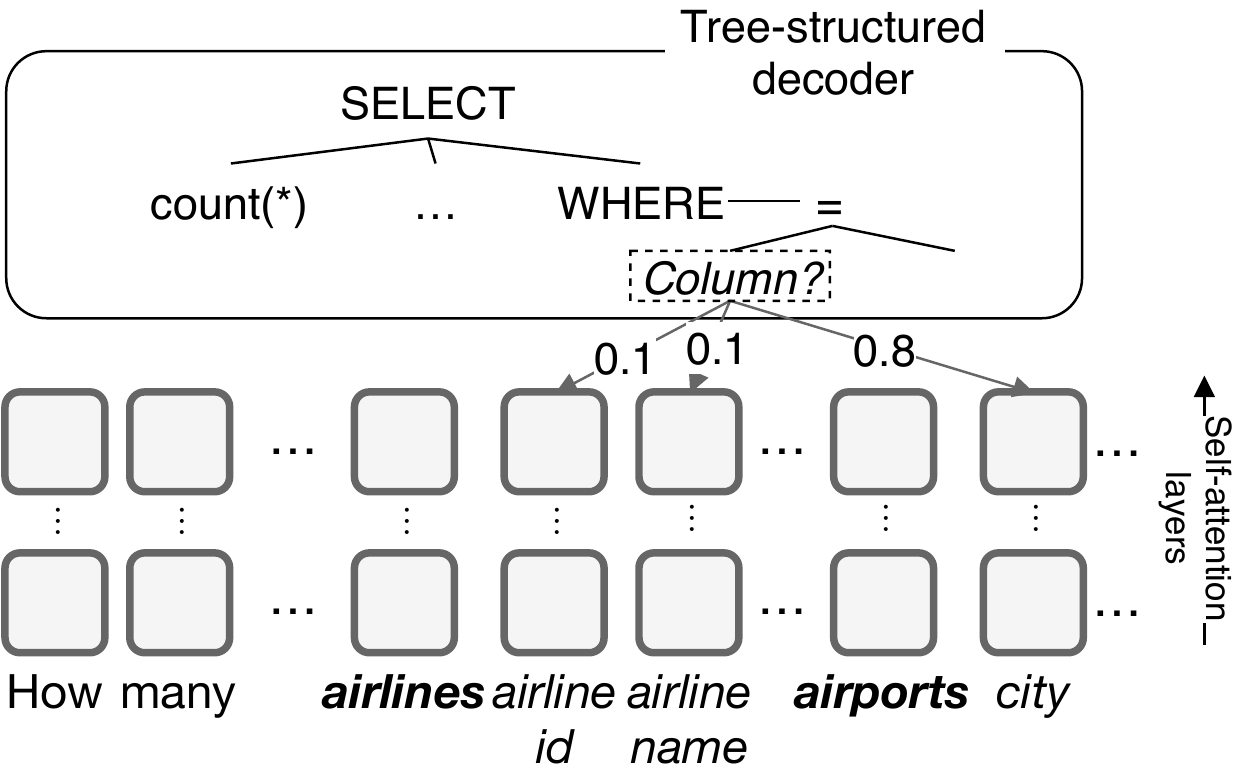}
        \caption{The decoder, choosing a column (Section~\ref{sec:decoder})}
        \label{fig:decoder}
    \end{subfigure}
    \caption{Overview of the stages of our approach.}\label{fig:approach}
\end{figure}

\subsection{Encoding the Schema as a Graph}
\label{sec:encoding-as-graph}
To support reasoning about relationships between schema elements in the encoder,
we begin by representing the database schema using a directed graph $\mathcal{G}$, where each node and edge has a label.
We represent each table and column in the schema as a node in this graph, labeled with the words in the name;
for columns, we prepend the type of the column to the label.
For each pair of nodes $x$ and $y$ in the graph, Table~\ref{table:schema-graph-edges} describes when there exists an edge from $x$ to $y$ and the label it should have.
Figure~\ref{fig:schema-graph} illustrates an example graph (although not all edges and labels are shown).

\subsection{Initial Encoding of the Input}
\label{sec:initial-encoding}
We now obtain an initial representation for each of the nodes in the graph, as well as for the words in the input question.
For the graph nodes, we use a bidirectional LSTM over the words contained in the label.
We concatenate the output of the initial and final time steps of this LSTM to form the embedding for the node.
For the question, we also use a bidirectional LSTM over the words.
Formally, we perform the following:
\begin{align*}
(\mathbf{c}_{i,0}^\text{fwd}, \mathbf{c}_{i,0}^\text{rev}), \cdots, (\mathbf{c}_{i,|c_i|}^\text{fwd}, \mathbf{c}_{i,|c_i|}^\text{rev})
& = \text{BiLSTM}_\text{Column}(c_{i}^\text{type}, c_{i,1}, \cdots, c_{i,|c_i|}); \quad
\mathbf{c}_{i}^\text{init} = \text{Concat}(\mathbf{c}_{i,|c_i|}^\text{fwd}, \mathbf{c}_{i,0}^\text{rev}) \\
(\mathbf{t}_{i,1}^\text{fwd}, \mathbf{t}_{i,1}^\text{rev}), \cdots, (\mathbf{t}_{i,|t_i|}^\text{fwd}, \mathbf{t}_{i,|t_i|}^\text{rev})
& = \text{BiLSTM}_\text{Table}(t_{i,1}, \cdots, t_{i,|t_i|}); \quad
\mathbf{t}_{i}^\text{init} = \text{Concat}(\mathbf{t}_{i,|c_i|}^\text{fwd}, \mathbf{t}_{i,1}^\text{rev}) \\
(\mathbf{q}_{1}^\text{fwd}, \mathbf{q}_{1}^\text{rev}), \cdots, (\mathbf{q}_{|q|}^\text{fwd}, \mathbf{q}_{|q|}^\text{rev})
& = \text{BiLSTM}_\text{Question}(q_{1}, \cdots, q_{|q|}); \quad
\mathbf{q}_{i}^\text{init} = \text{Concat}(\mathbf{q}_{i}^\text{fwd}, \mathbf{q}_{i}^\text{rev})
\end{align*}
where each of the BiLSTM functions first lookup word embeddings for each of the input tokens.
The LSTMs do not share any parameters with each other.

\subsection{Relation-Aware Self-Attention}
\label{sec:rel-attn}
At this point, we have representations $\mathbf{c}_{i}^\text{init}$, $\mathbf{t}_{i}^\text{init}$, and $\mathbf{q}_{i}^\text{init}$.
Similar to encoders used in some previous papers, these initial representations are independent of each other (uninfluenced by which other columns or tables are present).
Now, we would like to imbue these representations with the information in the schema graph.
We use a form of self-attention \citep{vaswaniAttentionAllYou2017} that is relation-aware \citep{shawSelfAttentionRelativePosition2018} to achieve this goal.

In one step of relation-aware self-attention, we begin with an input $\mathbf{x}$ of $n$ elements (where $x_i \in \mathbb{R}^{d_x}$) and transform each $x_i$ into $y_i \in \mathbb{R}^{d_z}$.
We follow the formulation described in \citet{shawSelfAttentionRelativePosition2018}:
\begin{align*}
    e_{ij}^{(h)} &= \frac{x_i W_Q^{(h)} (x_j W_K^{(h)} + \mathbf{\color{red}r_{ij}^K})^T}{\sqrt{d_z / H}}; \quad
    \alpha_{ij}^{(h)} = \frac{\exp(e_{ij}^{(h)})}{\sum_{l=1}^n \exp(e_{il}^{(h)})} \\
    z_i^{(h)} &= \sum_{j=1}^n \alpha_{ij}^{(h)} (x_j W_V^{(h)} + \mathbf{\color{red}r_{ij}^V}); \quad
    z_i = \text{Concat}(z_i^{(0)}, \cdots, z_i^{(H)}) \\
    \tilde{y}_i &= \text{LayerNorm}(x_i + z_i); \quad
    y_i = \text{LayerNorm}(\tilde{y}_i + \text{FC}(\text{ReLU}(\text{FC}(\tilde{y}_i)))
\end{align*}
The $r_{ij}$ terms encode the relationship between the two elements $x_i$ and $x_j$ in the input.
We explain how we obtain $r_{ij}$ in the next part.

\paragraph{Application Within Our Encoder}
At the start, we construct the input $x$ of $|c| + |t| + |q|$ elements using $\mathbf{c}_{i}^\text{init}$, $\mathbf{t}_{i}^\text{init}$, and $\mathbf{q}_{i}^\text{init}$:
\[
    x = (\mathbf{c}_{1}^\text{init}, \cdots, \mathbf{c}_{|c|}^\text{init},
         \mathbf{t}_{1}^\text{init}, \cdots, \mathbf{t}_{|t|}^\text{init}, 
         \mathbf{q}_{1}^\text{init}, \cdots, \mathbf{q}_{|q|}^\text{init}).
\]
We then apply a stack of $N$ relation-aware self-attention layers, where $N$ is a hyperparameter.
We set $d_z = d_x$ to facilitate this stacking.
The weights of the encoder layers are not tied; each layer has its own set of weights.

We define a discrete set of possible relation types, and map each type to an embedding to obtain $r_{ij}^V$ and $r_{ij}^K$.
We need a value of $r_{ij}$ for every pair of elements in $x$.
If $x_i$ and $x_j$ both correspond to nodes in $\mathcal{G}$ (i.e. each is either a column or table) with an edge from $x_i$ to $x_j$, then we use the label on that edge (possibilities listed in Table~\ref{table:schema-graph-edges}).

However, this is not sufficient to obtain $r_{ij}$ for every pair of $i$ and $j$.
In the graph we created for the schema, we have no nodes corresponding to the question words;
not every pair of nodes in the graph has an edge between them (the graph is not complete);
and we have no self-edges (for when $i = j$).
As such, we add more types beyond what is defined in Table~\ref{table:schema-graph-edges}:
\begin{itemize}
    \item If $i = j$, then \textsc{Column-Identity} or \textsc{Table-Identity}.
    \item $x_i \in $ question, $x_j \in $ question: 
       \textsc{Question-Dist-$d$}, where $d = \text{clip}(j - i, D)$; $\text{clip}(a, D) = \max(-D, \min(D, a))$. We use $D = 2$.
    \item $x_i \in $ question, $x_j \in \text{column} \cup \text{table}$; or $x_i \in \text{column} \cup \text{table}$, $x_j \in $ question: \\
       \textsc{Question-Column}, \textsc{Question-Table}, \textsc{Column-Question} or \textsc{Table-Question}
        depending on the type of $x_i$ and $x_j$.
    \item Otherwise, one of \textsc{Column-Column}, \textsc{Column-Table}, \textsc{Table-Column}, or \textsc{Table-Table}.
\end{itemize}
In the end, we add $2 + 5 + 4 + 4$ types beyond the $10$ in Table~\ref{table:schema-graph-edges}, for a total of 25 types.

After processing through the stack of $N$ encoder layers, we obtain
\begin{equation*}
     (\mathbf{c}_{1}^\text{final}, \cdots, \mathbf{c}_{|c|}^\text{final},
      \mathbf{t}_{1}^\text{final}, \cdots, \mathbf{t}_{|t|}^\text{final}, 
      \mathbf{q}_{1}^\text{final}, \cdots, \mathbf{q}_{|q|}^\text{final}) 
     = y.
\end{equation*}
We use $\mathbf{c}_{i}^\text{final}$, $\mathbf{t}_{i}^\text{final}$, and $\mathbf{q}_{i}^\text{final}$
in our decoder.

\paragraph{Comparison to Past Work}
We use the same formulation of relation-aware self-attention as \citet{shawSelfAttentionRelativePosition2018}.
However, that work only applied it to sequences of words in the context of machine translation, and as such, their $r_{ij}$ only encoded the relative distance between two words.
We extend their work and show that relation-aware self-attention can effectively encode more complex relationships that exist within an unordered sets of elements (in this case, columns and tables within a database schema).

Compared to the encoders used in past work such as Coarse2Fine \citep{dongCoarsetoFineDecodingNeural2018} and IncSQL \citep{shiIncSQLTrainingIncremental2018},
our novel use of relation-aware self-attention frees our encoder from spurious consideration of the order in which the columns and tables are presented in the schema (as the relations we have defined are not impacted by this order).

In their implementation, \citet{shawSelfAttentionRelativePosition2018} 
share $r_{ij}^K$ across the $H$ heads and the $b$ examples in a batch,
which meant they could use $n$ parallel multiplications of $bH \times (d_Z / H)$ and $(d_z / H) \times n$ matrices.
This is possible as $r_{ij}^K$ does not change across the batch when only encoding the relative distances between words.
However, due to the more varied relations between $x_i$ in our work which are not shared by all elements in a batch,
we instead use $bn$ parallel multiplications of $H \times (d_z / H)$ and $(d_z / H) \times n$ matrices,
exploiting the fact that we share $r_{ij}^K$ across the $H$ heads.

\subsection{Decoder}
\label{sec:decoder}
Once we have obtained an encoding of the input, we used the decoder from \citet{yinSyntacticNeuralModel2017a} to generate the SQL query.
The decoder generates the SQL query as an abstract syntax tree in depth-first traversal order, 
by outputting a sequence of \emph{production rules} that expand the last generated node in the tree.
However, following SyntaxSQLNet \citep{yuSyntaxSQLNetSyntaxTree2018}, the decoder does not generate the FROM clause; rather, it is recovered afterwards with hand-written rules using the columns referred to in the remainder of the query.
The decoder is restricted to choosing only syntactically valid production rules, and therefore it always produces syntactically valid outputs.
To save space, we refer readers to \citet{yinSyntacticNeuralModel2017a},
although we made the following modifications:
\begin{itemize}
    \item When the decoder needs to output a column, we use a pointer network based on scaled dot-product attention \citep{vaswaniAttentionAllYou2017} which points to $\mathbf{c}_{i}^\text{final}$ and $\mathbf{t}_{i}^\text{final}$.
    \item At each step, the decoder accesses the encoder outputs $\mathbf{c}_{i}^\text{final}$, $\mathbf{t}_{i}^\text{final}$, and $\mathbf{q}_{i}^\text{final}$ using multi-head attention. The original decoder in \citet{yinSyntacticNeuralModel2017a} uses a simpler form of attention.
\end{itemize}

\section{Experiments}
In this section, we describe the experiments we conducted to empirically validate our schema encoding approach.

\subsection{Experimental Setup}
We implemented our model using PyTorch \cite{paszkeAutomaticDifferentiationPyTorch2017}.
Within the encoder, we use GloVe word embeddings and hold them fixed during training.
All word embeddings have dimension $300$.
The bidirectional LSTMs have hidden size 128 per direction, and use the recurrent dropout method of \citet{galTheoreticallyGroundedApplication2016} with rate 0.2.
Within the relation-aware self-attention layers, we set $d_x = d_z = 256$, $H = 8$, and use dropout with rate $0.1$.
The position-wise feed-forward network has inner layer dimension 1024.
Inside the decoder, we use rule embeddings of size $128$, node type embeddings of size $64$, and a hidden size of $256$ inside the LSTM with dropout rate $0.2$ .

We used the Adam optimizer \citep{kingmaAdamMethodStochastic2014} with $\beta_1 = 0.9$, $\beta_2 = 0.999$, and $\epsilon = 10^{-9}$, which are defaults in PyTorch.
During the first $warmup\_steps = max\_steps / 20$ steps of training, we linearly increase the learning rate from 0 to $10^{-3}$.
Afterwards, the learning rate is annealed to 0, with formula $10^{-3}(1 - \frac{step - warmup\_steps}{max\_steps - warmup\_steps})^{-0.5}$.
For all parameters, we used the default initialization method in PyTorch.
We use a batch size of 50 and train for up to 40,000 steps.
\begin{table*}[t]
\centering

\caption{Exact match accuracy of different models on the development set of Spider.
The first row is the SyntaxSQLNet \citep{yuSyntaxSQLNetSyntaxTree2018} baseline;
the second row is our method;
the remainder are ablations on our method.
The columns refer to different subsets of the development set, from \citet{yuSpiderLargeScaleHumanLabeled2018};
``All'' refers to all 1,034 dev examples.
}
\label{table:exact-match}

\begin{tabular}{lrrrrr}
\toprule
Model & Easy & Medium & Hard & Extra Hard & All \\
\midrule
SyntaxSQLNet & 38.40\% & 15.00\% & 16.09\% & 3.53\%& 18.96\% \\
\midrule
Our method & 57.20\% & 44.55\% & 39.66\% & 21.18\% & \textbf{42.94}\% \\
\midrule
No self-attention layers & 42.00\% & 25.68\% & 22.99\% & 5.88\% & 25.92\% \\
2 self-attention layers & 56.00\% & 45.00\% & 40.23\% & 19.41\% & 42.65\% \\
\midrule
Fewer relation types & 47.60\% & 30.45\% & 25.86\% & 10.00\% & 30.46\% \\
No relation types & 46.80\% & 29.55\% & 29.89\% & 8.82\% & 30.37\% \\
\midrule
No pretrained word embeddings & 40.80\% & 29.09\% & 27.01\% & 5.88\% & 27.76\% \\
\bottomrule
\end{tabular}
\end{table*}

\subsection{Dataset and Metrics}
We use the Spider dataset \citep{data-spider} for all our experiments.
As described by \citet{data-spider}, the training data contains 8,659 examples, including 1,659 examples (questions and queries, with the accompanying schemas) from the Restaurants \citep{data-restaurants-original,data-restaurants-logic}, GeoQuery \citep{data-geography-original}, Scholar \citep{data-atis-geography-scholar}, Academic \citep{data-academic}, Yelp and IMDB \citep{data-sql-imdb-yelp} datasets.
We do \textbf{not} use the data augmentation scheme of \citet{yuSyntaxSQLNetSyntaxTree2018}.

As \citet{data-spider} have kept the test set secret, we perform all evaluations using the publicly available development set.
There are 1,034 examples in the development set, containing schemas distinct from those in the training set.
We report results using the same metrics as \citet{yuSyntaxSQLNetSyntaxTree2018}:
exact match accuracy on all development set examples, as well as after division into four levels of difficulty.
As in previous work, these metrics do not measure the model's performance on generating values within the queries.
We report results from the snapshot that obtained the best exact match accuracy across 3 repetitions of each configuration, except for the SyntaxSQLNet baseline where we reuse the pretrained model from the authors.

In addition to results on all of the development set, we also report results on subsets (Easy, Medium, Hard, and Extra Hard) partitioned by complexity of the query as defined by \citet{yuSpiderLargeScaleHumanLabeled2018}.
These partitions make up
24.18\%,
42.55\%,
16.83\%, and
16.44\% respectively.

\subsection{Variants Tested}
Our main result uses the encoder and decoder described previously, with the number $N$ of relation-aware self-attention layers in the encoder set to $4$.
To further study the utility of our scheme, we also tried the following variations, listed in Table~\ref{table:exact-match}.

\paragraph{Reducing the number of self-attention layers.}
Set $N = 0$ and $N = 2$. With $N = 0$, there are no relation-aware self-attention layers; we set
$\mathbf{c}_{i}^\text{final} = \mathbf{c}_{i}^\text{init}$,
$\mathbf{t}_{i}^\text{final} = \mathbf{t}_{i}^\text{init}$, and
$\mathbf{q}_{i}^\text{final} = \mathbf{q}_{i}^\text{init}$.
As such, the question words, the words in each column's name, and the words in each table's name are encoded separately using bidirectional LSTMs.

\paragraph{Removing the relation information from the encoder.}
We would like to measure the impact of providing to the encoder the 25 relation types we defined earlier.
In particular, we want to see whether the self-attention mechanism is sufficient within the encoder to obtain a representation for each schema element that is aware of all of the other schema elements, even if we don't explicitly provide information about how the elements are related.

For ``fewer relation types'', we exclude all of the types in Table~\ref{table:schema-graph-edges}, resulting in $15$ rather than $25$ possible types.
For ``minimal relation types'', we further merge all of \{\textsc{Question,Column,Table}\}-\{\textsc{Question,Column,Table}\} relations into one, as well as \{\textsc{Column,Table}\}-\textsc{Identity} with \textsc{Question-Dist-$0$}, and so we only have $6$ types.

\paragraph{Not using pretrained word embeddings.}
The Spider dataset only contains 8,659 training examples, which is significantly smaller than many other datasets used in natural language processing.
Furthermore, there is also reduced overlap in the vocabulary between the training and validation/test sets, as they contain different database schemas and domains.
Therefore, we measure the impact of using word embeddings learned from only this dataset.
We construct a vocabulary consisting of all of the words in the columns, tables, and questions that occur at least 3 times in the training data (the words which occur in columns and tables are counted every time the corresponding schema is used by a question in the data);
for each, we randomly initialize an embedding of dimension 300.

\section{Results and Discussion}

Table~\ref{table:exact-match} presents our exact match accuracy results on the development set of Spider. %
For the SyntaxSQLNet row, we obtained the results by running the pretrained model without data augmentation from \url{https://github.com/taoyds/syntaxSQL}.
Our method exceeds the performance of all other configurations tried, including all ablations.
In particular, we can see that our method strongly outperforms SyntaxSQLNet~\citep{yuSyntaxSQLNetSyntaxTree2018}, the best published baseline, achieving \bestresult{} exact match accuracy over the 18.96\% of the previous work.

\paragraph{Reducing the number of self-attention layers.}
We can see that the process of relation-aware self-attention is critical for the performance of this encoder, as the accuracy drops precipitiously when the self-attention layers are removed.
In particular, ``no self-attention layers'' uses an encoder very similar to SyntaxSQLNet's.
We observe fairly marginal gains by using $4$ such layers (in ``Our method'') as opposed to $2$ (``2 self-attention layers'').

\paragraph{Removing the relation information from the encoder.}
Comparing against the rows of ``No self-attention layers'' and ``Our method'', we see that while having self-attention layers helps increase performance, it is the relation information provided to the encoder that is responsible for most of the gains.
The use of self-attention, on its own, contributes relatively little.

\paragraph{Not using pretrained word embeddings.}
Given the small size of the training data, we confirm that using pretrained word embeddings helps significantly.
When we evaluate ``No pretrained word embeddings'' on the subset of the development set where all question words have a learned embedding (i.e. no UNKs in the question; 239 out of 1034 examples), then the exact match accuracy recovers to 40.17\%; ``Our method'' achieves 50.21\% on this subset,
substantialy reducing the gap in accuracy.

\section{Conclusion}
This paper proposes the use of relation-aware self-attention \citep{shawSelfAttentionRelativePosition2018} when encoding a database schema and a natural language question for the purposes of synthesizing a SQL query.
We achieve significantly better results on the Spider dataset
than the best published result of \citet{yuSyntaxSQLNetSyntaxTree2018}.
Our ablation study confirms the importance of encoding relations directly in the self-attention mechanism.

\bibliography{text2sql,text2sql-datasets}
\bibliographystyle{plainnat}

\end{document}